\documentclass{article}
\usepackage{spconf,amsmath,graphicx}
\usepackage{multirow}
\usepackage{enumitem}
\setlist{nosep, leftmargin=14pt}

\usepackage{todonotes}

\usepackage{mwe} 

\title{Normative Modeling via Conditional Variational Autoencoder and Adversarial Learning to Identify Brain Dysfunction in Alzheimer's Disease}
\name{Xuetong Wang$^{1}$, Kanhao Zhao$^{2}$, Rong Zhou$^{1}$, Alex Leow$^{3}$, Ricardo Osorio$^{4}$, Yu Zhang$^{2,5}$, Lifang He$^{1}$}
\address{$^{1}$ Department of Computer Science and Engineering, Lehigh University, PA, USA\\
$^{2}$ Department of Bioengineering, Lehigh University, PA, USA\\
$^{3}$Department of Psychiatry, University of Illinois at Chicago, IL, USA\\
$^{4}$ Department of Psychiatry, NYU Grossman School of Medicine, NY, USA \\
$^{5}$ Department of Electrical and Computer Engineering, Lehigh University, PA, USA}

\begin{document}
%
\maketitle
\begin{abstract}
Normative modeling is an emerging and promising approach to effectively study disorder heterogeneity in individual participants. 
In this study, we propose a novel normative modeling method by combining conditional variational autoencoder with adversarial learning (ACVAE) to identify brain dysfunction in Alzheimer's Disease (AD). Specifically, we first train a conditional VAE on the healthy control (HC) group to create a normative model conditioned on covariates like age, gender and intracranial volume. Then we incorporate an adversarial training process to construct a discriminative feature space that can better generalize to unseen data. Finally, we compute deviations from the normal criterion at the patient level to determine which brain regions were associated with AD. Our experiments on OASIS-3 database show that the deviation maps generated by our model exhibit higher sensitivity to AD compared to other deep normative models, and are able to better identify differences between the AD and HC groups.


\end{abstract}
\section{Introduction}
\label{sec:intro}
Brain diseases such as Alzheimer's disease (AD) are usually highly heterogeneous, which exacerbates the difficulty of clinical treatment. The traditional case-control approaches assume a consistent pattern of abnormalities among individuals belonging to the same cohort, but ignore the heterogeneity of the disorder \cite{li1992case, badhwar2020multiomics}. In contrast to case-control studies, normative modeling can quantify how individual patients deviate from the expected normative range by learning in the healthy control (HC) group and thus explicitly modeling disease heterogeneity, which provides information about potential abnormalities in each particular individual. The normalized model is a two-step process in which the model is first trained on the HC cohort, and then the trained model is applied to a target cohort to quantify deviations \cite{mohammadian2018novelty}.


Recently, deep learning techniques are very popular for normative modeling, especially autoencoder (AE) based methods \cite{chamberland2021detecting, pinaya2019using}. AE consists of two components: encoder and decoder. The encoder compresses the data from a high-dimensional space to a low-dimensional space also called latent code, and then the decoder converts the data from the latent code to a high-dimensional space like the input data. However, since AE mainly emphasizes the image reproduction function, this also leads to a drawback that is the lack of randomness, which results in a model that is not sufficiently generalized. When we randomly change the latent code, the output may not be related to the original data at all. In other words, the latent code for AE is non-regularized. 

To solve the above issue, variational autoencoder (VAE) \cite{lawry2022conditional} is adopted in the normative modeling. VAE is a generative model that differs from the AE in that the encoder of VAE outputs the parameters of a pre-defined distribution rather than just a latent vector. It then forces this distribution to be a normal distribution, which ensures that the latent space is regularized. Moreover, to eliminate the influence of some confounding variables (e.g., age, gender, and intracranial volume) on the model in the learning process, conditional variational autoencoder (CVAE) has also been applied to the normative modeling \cite{lawry2022conditional}. Unlike the vanilla VAE, the encoder of CVAE can generate latent distribution parameters defined in advance based on the input data and confounding variables. Similarly, the decoder can reconstruct the original input based on the confounding variables and the vectors sampled in the latent space. On the other hand, there are some recent studies applying adversarial learning to autoencoders (AAE) in normative modeling \cite{pinaya2021using, kim2021conditional}, but the learning does not sufficiently take into account the randomness in the latent space which may lead to unreliable result in the analysis.

\begin{figure*}[t]
\centerline{\includegraphics[width=0.7\linewidth]{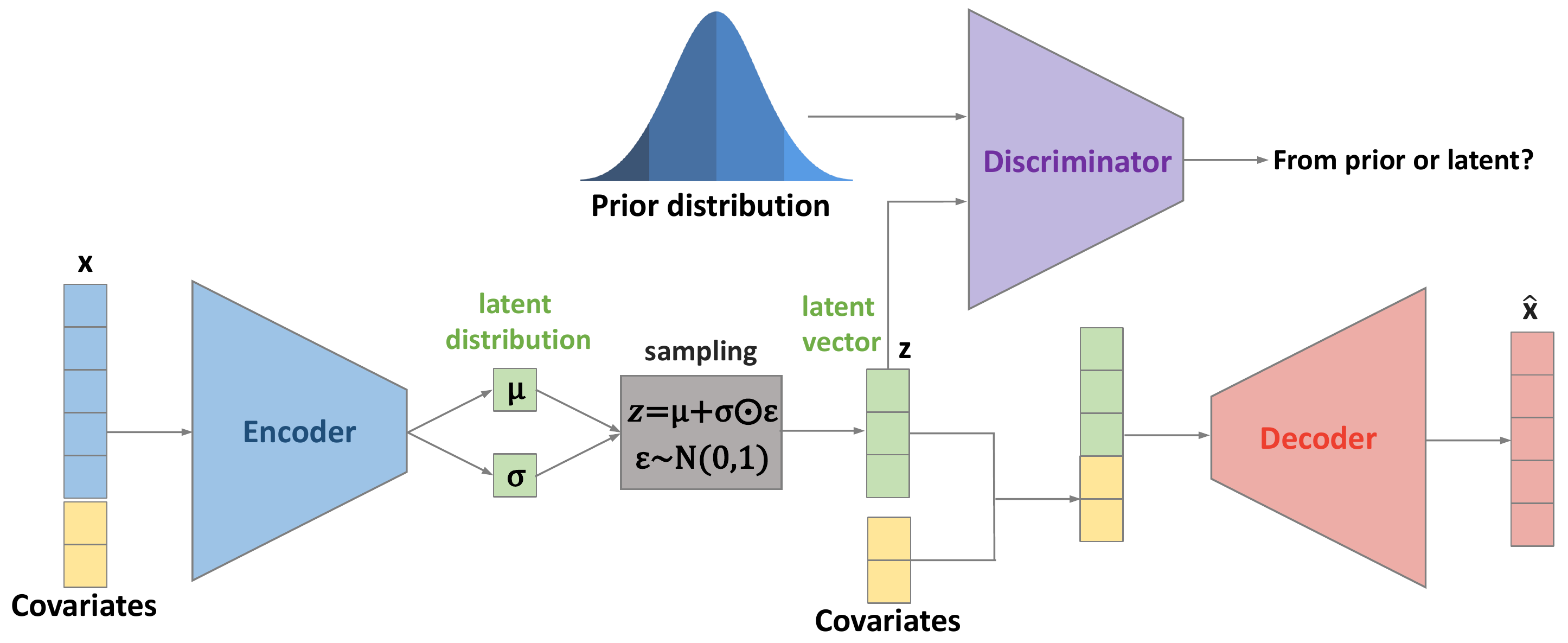}}
\caption{The framework of the adversarial conditional variational autoencoder.}
\label{fig1}
\vspace{-5pt}
\end{figure*}

In this paper, we propose a novel normative model named as ACVAE by integrating CVAE and adversarial learning that improves the generality of the model by adding randomness to the latent space and effectively reduces the impact on model learning by passing confounding variables in the encoder and decoder. First, we used regional brain volume data from the HC group to train our model. Next, we tested the trained model in the target population, generating deviation maps for AD and HC groups. As a result, each patient's deviation from the norm was estimated. By comparison with other models, our model exhibited better separation in terms of deviation between HC and AD subjects. By observing different deviation plots for each patient, heterogeneity can be better understood, which can provide a more reliable reference basis for clinical diagnosis and treatment.

\section{Materials and Methods}
\label{sec:method}
\subsection{fMRI Acquisition and Preprocessing}
The fMRI data used in this study was obtained from the OASIS-3 database \cite{lamontagne2019oasis}. It included a total of 1098 subjects with 605 cognitively normal adults and 493 individuals at various stages of cognitive decline ranging in age from 42 to 95 years. For each session, the fMRI data were scanned during resting state for 6 min (164 volumes) using 16-channel head coil of scanners with parameters: TR$=$2.2 s, TE$=$27 ms, FOV$=$240$\times$240 mm, and FA$=$90$^\circ$. The acquired rs-fMRI data were preprocessed using the reproducible fMRIPrep pipeline \cite{esteban2019fmriprep}. The T1-weighted image was corrected for intensity nonuniformity and then stripped skull as T1w-reference. Spatial normalization was done through nonlinear registration, with the T1w reference \cite{avants2008symmetric}, followed by FSL-based segmentation. The BOLD reference was then transformed to the T1w reference with a boundary-based registration method, configured with nine degrees of freedom to account for distortion remaining in the BOLD reference \cite{greve2009accurate}. BOLD signals were slice-time corrected and resampled onto the participant's original space with head-motion parameters \cite{jenkinson2002improved}, susceptibility distortion correction, and then resampled into standard space, generating a preprocessed BOLD run in MNI standard space. ICA-AROMA \cite{pruim2015ica} was then performed for automatic removal of motion artifacts.

\begin{table}[t]
\small
\centering
\caption{Data Characteristics.} \label{table_data}
\begin{tabular}{l|l|l}
\hline
                               & HC                               & AD                               \\\hline
Num                            & 1476                             & 21                               \\
Gender (M/F)                    & 647/829                          & 11/10                            \\
Age (mean$\pm$ std) & 68.5$\pm$6.4                         & 77.4$\pm$3.4        \\
ICV (mean$\pm$ std) & 60960.2$\pm$10249.9 & 61172.8$\pm$10493.2 \\ \hline
\end{tabular}
\vspace{-10pt}
\end{table}

\subsection{Feature Generation}
Table \ref{table_data} shows the characteristics of healthy control (HC) and Alzheimer's disease (AD) subjects used in this study. The OASIS-3 database spans a considerable amount of collection time, and some subjects were collected from multiple time periods. Thus, we treated the data of every 100-day period as a subject and assumed no significant deviation change in the same subject during the 100-day interval. This will result in 1497 subjects with 1476 HC subjects and 21 AD patients. For each subject, the voxel-level BOLD time series were first averaged into 100 regions-of-interest (ROIs) at each time point based on the Schaefer parcellation \cite{schaefer2018local}, and then averaged across time points to generate the ROI features as input. In addition, we consider the age, gender, and intracranial volume (ICV) as potentially confounding covariates, which were included in our model as conditional variables. The ICV for each subject was calculated by summing together all 100 ROI values. Both age and ICV were divided into 10 quantile-based bins and featured as one-hot encoded vectors. Specifically, the covariates from each subject were represented as a 22-dimensional vector with two dimensions used to one-hot encode the gender of male and female attributes.

\vspace{-5pt}
\subsection{Normative Modeling}
\textbf{Overview.} Fig.~\ref{fig1} illustrates our end-to-end architecture of ACVAE for normative modeling consisting of two components: conditional variational autoencoder (bottom) and adversarial learning (top). The former generates useful low-dimensional latent representations of ROI features in HC group and is conditioned by confounding covariates, while the latter employs adversarial training to shape the latent code distribution so that it resembles the predetermined prior distribution. Next, we give the details of each component.





\textbf{Conditional Variational Autoencoder (CVAE).}
CVAE is a variant of variational autoencoder (VAE) to conditional tasks, which allows to learn a low-dimensional latent space from the input data with multivariable control in an unsupervised manner. To eliminate the effect of confounding covariates present in the data on the latent space in the neural network, we used a CVAE model for normative modeling. Similar to VAE, the CVAE network has three main components: the encoder, the latent distribution, and the decoder. However, the encoder and decoder of the CVAE receive additional conditional variables, i.e., age, gender, and ICV. First, the encoder receives the conditional variables and input data, and then generates pre-defined distribution parameters, i.e., mean and variance. The decoder receives both the samples sampled from the latent distribution and the conditional variables to output the reconstructed input. Furthermore, the loss function can be formulated as:
\begin{equation}
     L\textsubscript{CVAE} = E[logP(X|z,c)]-D\textsubscript{KL}[Q(z|X,c)||P(z|c)],
\end{equation}
where $X$ is the input data, $c$ is the conditional variables, and $z$ is the latent code. In addtion, $Q(z|X,c)$, $P(X|z,c)$, $P(z|c)$ are represented as encoder, decoder, prior respectively. The first term of the above loss function is the reconstruction mean squared error that measures the difference between the input data and the reconstructed output. The second term refers to the Kullback-Leibler (KL) divergence, which measures how far the pre-defined distribution ($Q(z|X,c)$) is from the true distribution ($P(z|c)$).

\textbf{Adversarial Learning.}
Previous study has shown that combining VAE with adversarial learning allows to complement the VAE reconstruction loss with the perceptual-level representation of the discriminator \cite{wang2018learning}. Thus we combine the benefit of adversarial learning with CVAE. Adversarial learning consists of two parts: the discriminator and the generator. The discriminator accepts two inputs, one is a random sample sampled in the prior distribution and the other is sampled from the latent distribution. The discriminator will try to identify whether the input is a sample sampled from the prior distribution or sampled from the latent distribution. Yet the generator wants to generate samples that cheat the discriminator. The objective function can be expressed as follows:
\begin{equation}
    L\textsubscript{Adv} = E[log(D(z))]+E[log(1-D(P(X|z,c)))],
\end{equation}
where $D(z)$ is the discriminator and $P(X|z,c))$ is the generator, in this case also the decoder. Since the discriminator will output smaller values for samples from the prior distribution, otherwise it will output higher values. As a result, the discriminator seeks to maximize the loss. However, the generator wants the discriminator to think that the generated data is a sample from a priori distribution, so the generator wants to minimize the loss. This constitutes an adversarial learning process between discriminator $D$ and generator $G$ to concurrently train their respective neural networks.

\textbf{Deviation Metric.}
Similar to the previous work \cite{pinaya2021using}, we used the standard mean square error (MSE) as a performance function to compute the deviation between the input data and the reconstructed output as follows:

\begin{equation}\label{equation:1}
    D\textsubscript{MSE} = \frac{1}{|R|}\sum_{i\in R}(x_i-\hat{x_i})^2,
\end{equation}
where $x_i$ is the value of the brain region $i$ after normalization, $\hat{x_i}$ is the value of the brain region $i$ reconstructed by the decoder. $R$ is a set representing all brain regions of interest and $|R|$ denotes the cardinality of $R$ (i.e., $|R| = 100$).




\begin{table}[t]
\small
\centering
\caption{Performance Comparison.} \label{table1}
\begin{tabular}{l|l|c}
\hline
Category                               & Method       & ROC-AUC \\ \hline
\multirow{2}{*}{Non-Conditional Model} & AE           & 60.41\%   \\ \cline{2-3} 
                                       & VAE          & 62.66\%  \\ \hline
\multirow{3}{*}{Conditional Model}     & AAE          & 63.88\%   \\ \cline{2-3} 
                                   & CVAE         & 64.67\%   \\ \cline{2-3} 
                                       & ACVAE (ours) & 66.25\%   \\ \hline
\end{tabular}
\vspace{-15pt}
\end{table}


\section{Experiments}

\textbf{Experimental Settings.}
We divided the whole data into a training set and a test set. The training set was obtained by randomly selecting 80\% of the HC subjects, and the rest was grouped into the test set along with the AD patients. Specifically, we scaled the ROI values of each subject by dividing them by the ICV (as a means of adjusting for different brain sizes). Then, we normalized the training and test sets separately using the robust scaler method from the scikit-learn library, which is robust to outliers. We first scaled the features of the training set by subtracting the median and then dividing by the interquartile range (25\% value -- 75\% value). After this, we used the same statistics (median and interquartile range) of training set to normalize the test set. In order to ensure the robustness of the results, we used the bootstrap resampling technique to repeat the operation 10 times and reported the averaged results.


\textbf{Competing Methods.}
To demonstrate the performance of the proposed ACVAE, we compared it with four other deep normative modeling methods: the vanilla AE \cite{chamberland2021detecting}, VAE \cite{lawry2022conditional}, CVAE \cite{lawry2022conditional}, and AAE \cite{pinaya2020normative}, each of which represents a different normative strategy. In particular, both CVAE and AAE are conditional autoencoder methods, and we also used age, gender, and ICV as conditional variables of these models to have a fair comparison.

\begin{figure*}[htpb]
\centerline{\includegraphics[width=1\linewidth]{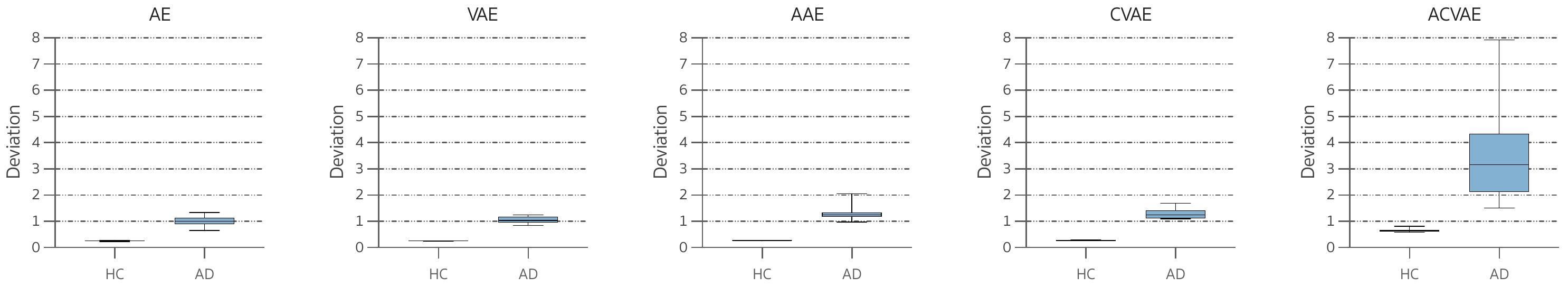}}
\caption{The observed mean deviation of HC and AD for each method. }
\label{fig:2}
\vspace{-8pt}
\end{figure*}

\begin{figure*}[htpb]
\centerline{\includegraphics[width=1\linewidth]{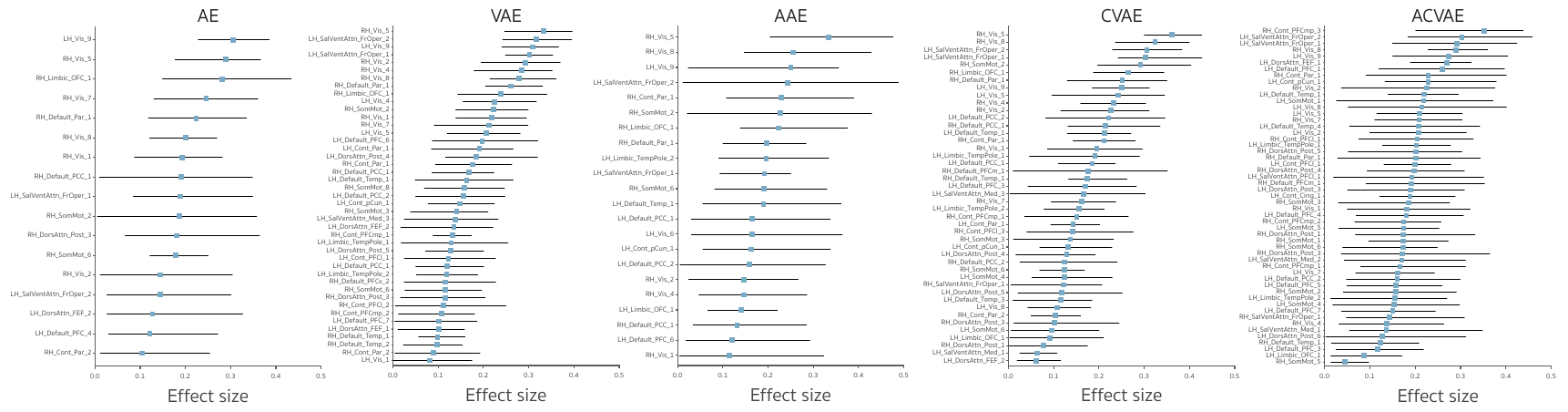}}
\caption{Mean effect sizes for HC and AD. The $y$-axis depicts brain regions selected for each method with significant differences.}
\label{fig:3}
\vspace{-8pt}
\end{figure*}

\textbf{Parameter Screening.}
In the proposed ACVAE, both the encoder and decoder architectures are highly flexible in code length and code rate, here we used two hidden layers for each module. We filtered different potential latent dimensions (10, 20) and encoder/decoder layer sizes (90, 100, and 110). We used the Adam optimizer for training with a total of 200 epochs and a cyclic learning rate containing a minimum bound and a maximum bound \cite{smith2017cyclical}, where the minimum bound is set to 0.0001 and the maximum bound is set to 0.005. The decay parameter gamma was set to 0.98. Finally, the mini-batch method with a batch size of 256 was used for training. In addition, to avoid the gradient vanishing problem, we used LeakyReLU as our activation function. The weight coefficient of $L\textsubscript{CVAE}$ loss was chosen among \{0.1, 1, 5, 10, 100, 1000\}, and the weight coefficient of the discriminator loss $L\textsubscript{Adv}$ was chosen among \{0, 2, 4, ..., 200\}. For other competing methods, we use their public codes and the same parameter settings for a fair comparison.

\textbf{Results.}
Table \ref{table1} shows the performance comparison of five methods in terms of the ROC-AUC score. From the experimental results, it can be observed that the proposed method outperforms the other competing methods, especially our model improves from 64.67\%  to 66.25\% on the basis of CVAE, which is the best baseline method. Additionally, we observed that the method that takes conditional variables into account is superior to the method that does not. This demonstrated the ability of ACVAE to separate the influence of covariates from the latent vectors. Moreover, Fig.~\ref{fig:2} shows the deviation boxplots for the HC and AD groups for each model in the test set, which is the corresponding deviation average over 10 repeated experiments. It can be seen that our ACVAE model can better distinguish HC and AD. Furthermore, we explore which brain regions had a larger effect. To this end, we calculated 95\% confidence intervals for the effect size of the difference in mean deviations for HC and AD. If the interval contains 0, it means that the difference is insignificant, otherwise, it means that the difference is significant. Fig. \ref{fig:3} shows the selected brain regions of each method with the most significant differences. Compared to the baselines, our model is capable of detecting more brain regions with significant effects. Taken together, this suggests that our method is more robust and sensitive.

\vspace{-13pt}

\section{Conclusion}
In this paper, a new normative model is presented for quantifying deviations in Alzheimer's disease (AD) and health control (HC) at the individual level, named as adversarial conditional variational autoencoder (ACVAE). Unlike the case-control studies, ACVAE does not require training in a dataset with a reasonable balance of AD and HC groups. It is trained with only HCs, allowing it to use large cohorts of HC participants. Besides, it enables to reduce the effect of covariates in the external cohort, which highlights the potential for covariate adjustment. We validated ACVAE on OASIS-3 dataset and the experimental results showed that our approach is more effective to identify deviation values than existing models, demonstrating its potential for identifying minor pathogenic effects. Because this kind of disorder is associated with profound changes in brain morphology that are not present in the training set, this pattern is expected. This means that our approach could be applied to develop more reliable and personalized treatment plans for a variety of patients.

\section{Compliance with Ethical Standards}
This research study was conducted retrospectively using human subject data made available in open access by OASIS-3 database \cite{lamontagne2019oasis}. Ethical approval was not required as confirmed by the license attached with the open access data.

\vspace{5pt}





\bibliographystyle{IEEEbib}
\bibliography{reference}

\end{document}